\documentclass[letterpaper, 10 pt, conference]{ieeeconf}  

\IEEEoverridecommandlockouts                              

\usepackage{graphicx}
\usepackage{amsfonts}
\usepackage{stfloats}
\usepackage{amsmath}
\usepackage{multirow}
\usepackage{float} 
\usepackage{booktabs}
\usepackage{tabularx}
\usepackage[dvipsnames]{xcolor}
\usepackage{threeparttable}
\usepackage{ragged2e}
\usepackage{cite}

\title{\LARGE \bf
Cross-modal State Space Modeling for Real-time RGB-thermal Wild Scene Semantic Segmentation
}

\author{Xiaodong Guo, Zi'ang Lin, Luwen Hu, Zhihong Deng, Tong Liu, and Wujie Zhou, Senior \emph{Member, IEEE}
\thanks{*This work was supported by National Natural Science Foundation of China under Grant 62476026. (\emph{Corresponding author: Tong Liu}).}
\thanks{Xiaodong Guo, Zi'ang Lin, Luwen Hu, Zhihong Deng and Tong Liu are with the School of Automation, Beijing Institute of Technology, Beijing 100081, China (e-mail: 
        {\tt\small liutong2002@bit.edu.cn, 3120245534@bit.edu.cn}).}%
\thanks{Wujie Zhou is with the School of Information and Electronic Engineering, Zhejiang University of Science and Technology, Hangzhou 310023, China, and also with the School of Computer Science and Engineering, Nanyang Technological University, Singapore 308232.}%
}

\begin{document}

\maketitle
\thispagestyle{empty}
\pagestyle{empty}

\begin{abstract}

The integration of RGB and thermal data can significantly improve semantic segmentation performance in wild environments for field robots. Nevertheless, multi-source data processing (e.g. Transformer-based approaches) imposes significant computational overhead, presenting challenges for resource-constrained systems. To resolve this critical limitation, we introduced CM-SSM, an efficient RGB-thermal semantic segmentation architecture leveraging a cross-modal state space modeling (SSM) approach. Our framework comprises two key components. First, we introduced a cross-modal 2D-selective-scan (CM-SS2D) module to establish SSM between RGB and thermal modalities, which constructs cross-modal visual sequences and derives hidden state representations of one modality from the other. Second, we developed a cross-modal state space association (CM-SSA) module that effectively integrates global associations from CM-SS2D with local spatial features extracted through convolutional operations. In contrast with Transformer-based approaches, CM-SSM achieves linear computational complexity with respect to image resolution. Experimental results show that CM-SSM achieves state-of-the-art performance on the CART dataset with fewer parameters and lower computational cost. Further experiments on the PST900 dataset demonstrate its generalizability. Codes are available at https://github.com/xiaodonguo/CMSSM.

\end{abstract}

\section{INTRODUCTION}

Accurate semantic segmentation of the surrounding environment is fundamental to the automation of field robots working in wild scenes. In addition to the color and texture information captured by Red-Green-Blue (RGB) cameras, the thermal radiation of targets provides valuable information, which can be obtained using long-wave thermal cameras. For instance, differences in thermal radiation can help distinguish shrubs and the bare ground when their colors are similar\cite{1}.  Therefore, integrating RGB-thermal (RGB-T) information is a valuable research direction for enhancing semantic segmentation in wild scenes. An effective and efficient cross-modal feature fusion (CFF) strategy is the key to accurate RGB-T semantic segmentation.
\begin{figure}[thpb]
      \centering
      \includegraphics[scale=0.8]{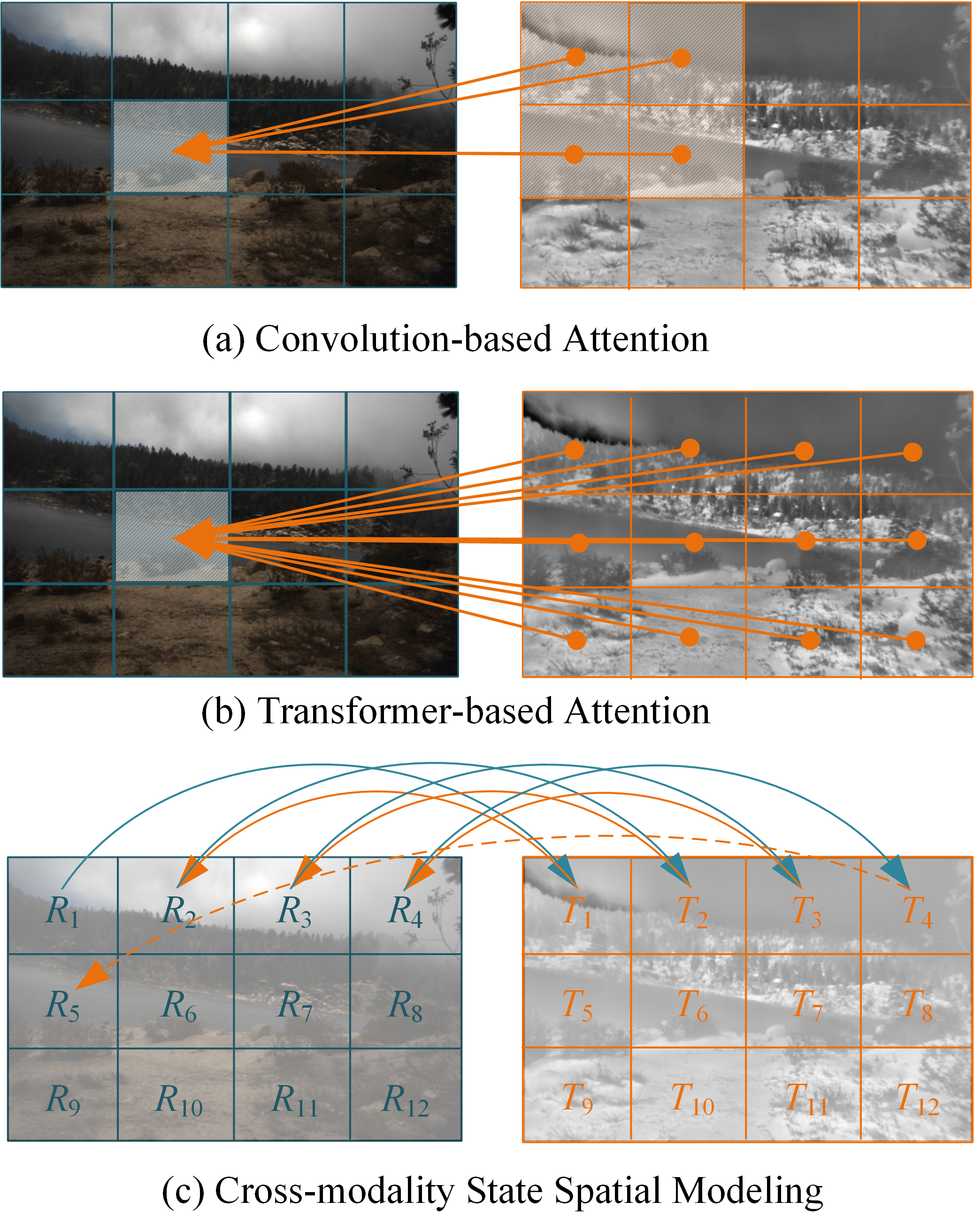}
      \caption{Three cross-modal feature fusion strategies: (a) convolution-based attention, (b) Transformer-based attention and (c) cross-modality state spatial modeling.}
      \label{figure1}
\end{figure}

The primary strategies for CFF were simple but insufficient, such as directly adding or multiplying RGB-T features. Although these methods did not introduce additional parameters and imposed a small inference burden, they ignored the intrinsic differences between the RGB-T information. Further, convolution-based attention methods have been introduced to adjust the fusion weights of the two modalities, as shown in Fig. 1 (a). The fusion weights can be obtained along the spatial scale and adaptively adjusted through back-propagation, which emphasize the effective information in each modality \cite{2, 3, 35}. However, the limited size of convolution kernel size cannot capture spatially distant information. One solution is to increase the size of the convolution kernel; while the significant increase in inference burden and parameter count renders it impractical. Another approach is to leverage the self-attention mechanism \cite{4} to establish long-distant dependency, which we refer to as Transformer-based attention. As shown in Fig. 1 (b), similarity relationship of each
\begin{figure*}[t]
      \centering
      \includegraphics[width=\textwidth]{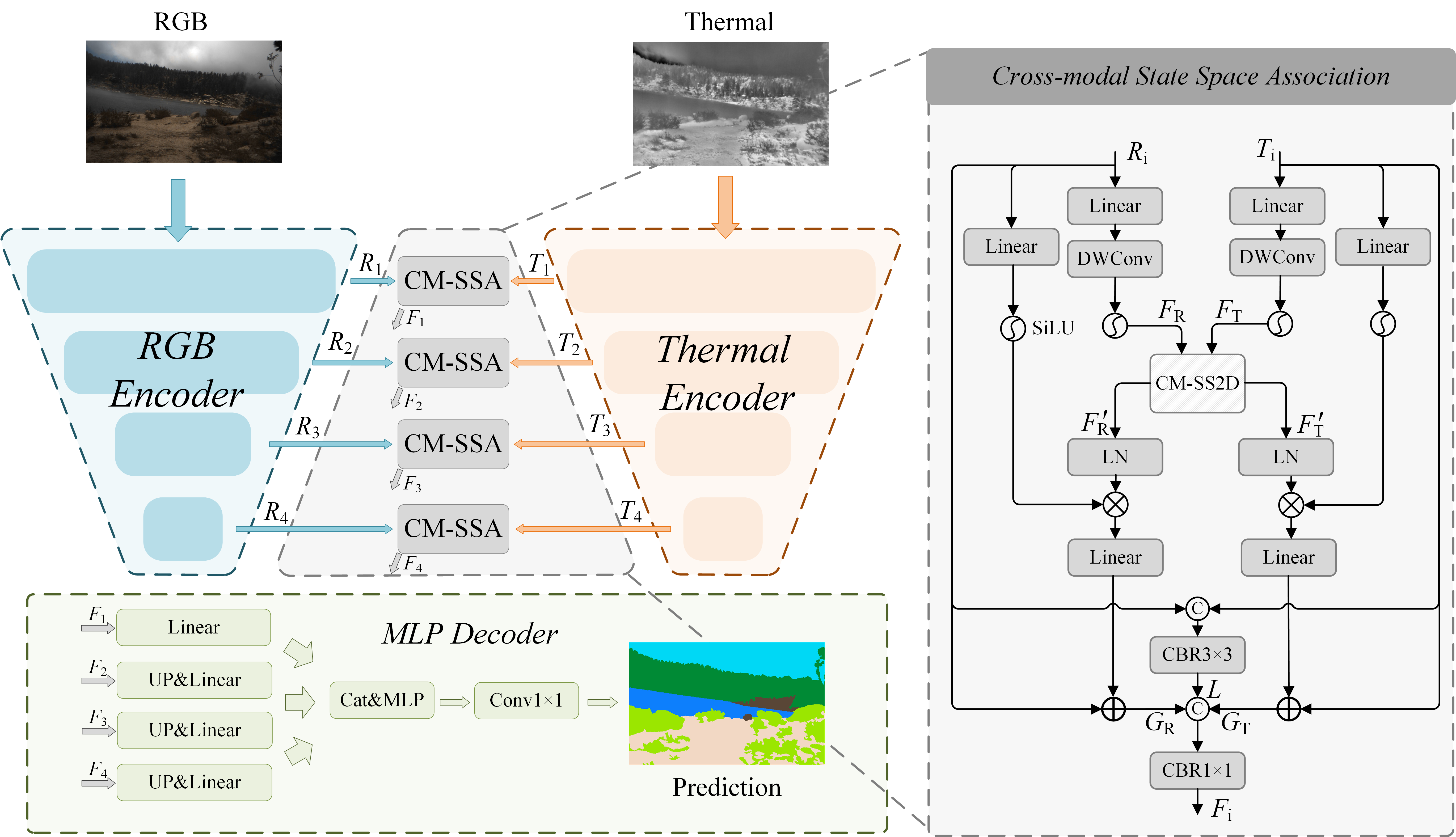}
      \caption{Illustration of CM-SSM. CM-SSM consists of two image encoders to extract the feature of RGB and thermal images, four CM-SSA modules to perform RGB-T feature fusion in four stages, and an MLP decoder to predict the semantic segmentation maps.}
      \label{figure2}
\end{figure*}
pixel in the two modalities are established to build global attention association \cite{36,8}. However, the computational complexity of the self-attention mechanism is quadratic with respect to image resolution. When processing high-resolution images, this increased computational burden leads to a significant decrease in inference speed. While Zhang et al. \cite{5, 6} leveraged channel-wise Transformer to alleviate computational burden, this approach inevitably weakens the model’s capacity to model global spatial relationships.

Recently, state spatial model (SSM) was applied in natural language processing \cite{9,10,11,12} and computer vision tasks \cite{14,15}, aiming to reduce the complexity of the model when building long-distant dependency of the feature sequence. Specifically, each output of the SSM is determined by the current input and the hidden state. The hidden state is obtained by coupling the current 
input with the previous state. In this way, the hidden state gradually incorporates information from the entire sequence, while the computational complexity of the output is reduced to linear with respect to the sequence length. Based on this, Liu et al. \cite{15} proposed VMamba, with a 2D-Selective-Scan (SS2D) module to build vision sequence from four directions. Subsequently, the SSM was modified for CFF in three ways, including 1) performing feature enhancement for each modality separately \cite{16}, 2) improving the spatial scanning strategy \cite{17} and 3) obtaining the state space parameters of one modality from the other \cite{18,34}. However, the above methods ignore the core relationship between SSM and CFF, namely how to construct the hidden state across different modalities. 

In contrast with previous studies, we proposed a real-time RGB-T semantic segmentation model based on cross-modal SSM, called CM-SSM. The CM-SSM includes a cross-modal 2D-selective-scan (CM-SS2D) module, which builds an SSM across RGB and thermal features. As shown in Fig. 1 (c), CM-SS2D constructs a cross-modal vision feature sequence through ‘RGB–thermal–RGB’ scanning. Subsequently, it obtains the hidden state by combining the input of one modality and the state of the other. In this way, the hidden state contains information from the same position of the other modality and gradually covers the entire global context. In addition, we proposed a cross-modal state space association (CM-SSA) module to combine the global association from CM-SS2D and local association from convolution.

Our main contributions can be summarized as follows:
\begin{itemize}
\item We proposed a real-time RGB-T semantic segmentation model, CM-SSM, for field robots working in wild scenes. It has 12.59 M parameters (Params) and requires 10.34 G floating-point operations (FLOPs), achieving an inference speed of 114 frames per second (FPS) on an RTX 4090 GPU.
\item We proposed the CM-SS2D and CM-SSA modules to achieve multi-modal feature fusion. The CM-SS2D module constructs cross-modal vision sequence and obtains the hidden state of one modality through the other. The CM-SSA module combines the cross-modal global and local association. 
\item We evaluated the performance of CM-SSM on the CART and PST900 datasets and demonstrated its superiority over state-of-the-art (SOTA) methods.

\end{itemize}
\begin{figure*}[t]
      \centering
      \includegraphics[width=\textwidth]{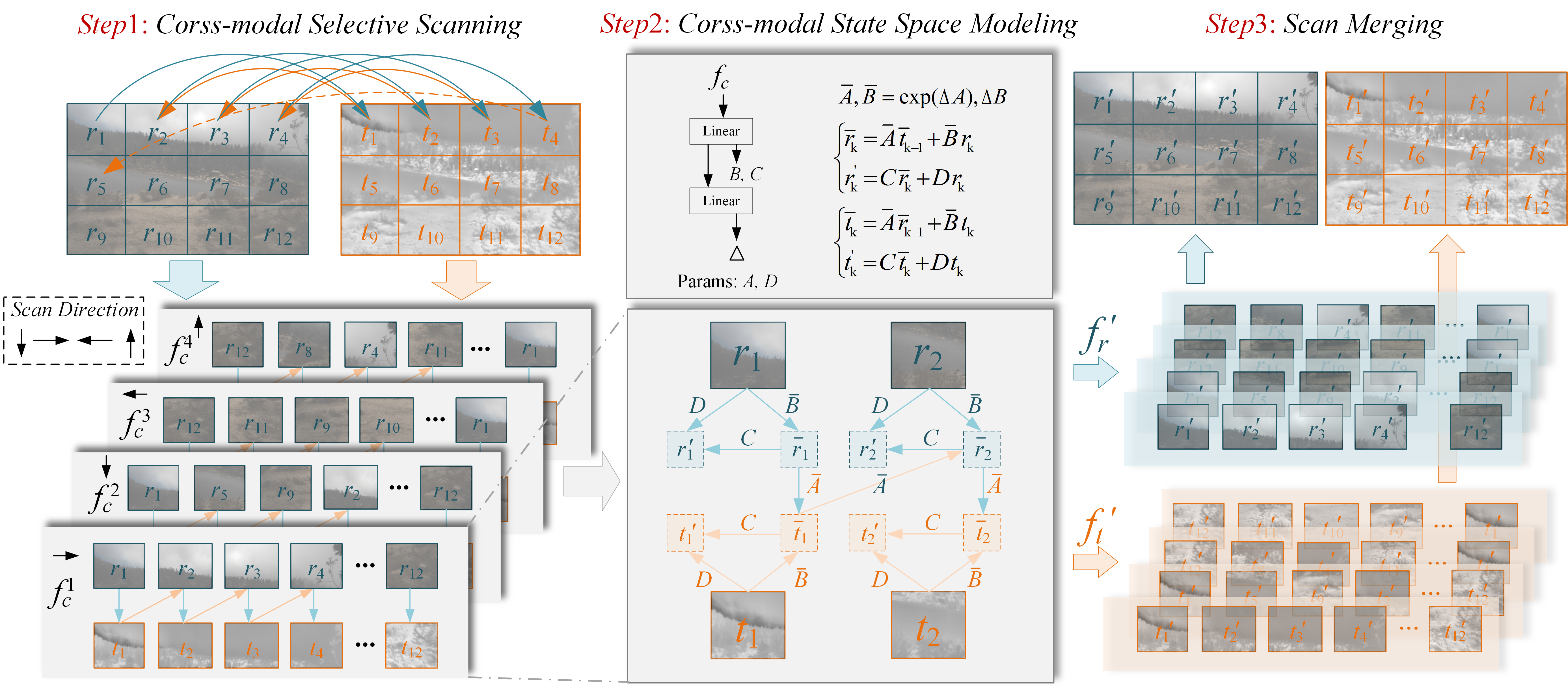}
      \caption{Illustration of CM-SS2D. CM-SS2D consists of three steps: 1) cross-modal selective scanning, 2) cross-modal state space modeling, and 3) scan merging.}
      \label{figure3}
\end{figure*}

\section{RELATED WORK}

\subsection{RGB-T Semantic Segmentation}

RGB-T semantic segmentation aims to leverage both RGB and thermal information to improve the segmentation accuracy of models under adverse conditions. The key to improving model accuracy lies in efficiently utilizing the complementary nature of cross-modal information. A simple approach is to concatenate the RGB-T features \cite{19,20}, ignoring the inherent differences between the features. Subsequently, feature fusion methods based on spatial and channel attention were proposed to highlight locally significant regions. GMNet \cite{3} and CLNet-T \cite{2} utilized the spatial attention at low-dimensional features and channel attention at high-dimensional features, according to the different characteristics of the features. To establish global relationships, Transformer \cite{4} was introduced to establish global relationships. MFTNet \cite{7}, MCNet-T \cite{36} leveraged features from various modalities to generate query, key, and value vectors, enable multi-modal feature interaction. Although accuracy improves, the Transformer imposes a 
significant computational burden, making it impractical for resource-constrained field robots. In contrast, we proposed the CM-SSM based on SSM. It establishes cross-modal global correlations while imposing a lower computational burden.

\subsection{State Space Model}

SSM is a fundamental concept in control theory, describing the relationship between input, output, and state variables. LLSL \cite{9}, S4 \cite{10} extended it into deep learning with derivations. Subsequently, S5 \cite{12} and H3 \cite{11} further explored the potential of SSM. Based on the former research, Mamba \cite{13} propose a selection mechanism for dynamic feature extraction. The mamba outperformed transformers based models across various 1D tasks with fewer computational resources occupied. Subsequently, VMamba \cite{15} developed the SSM for 2D vision tasks, with the assistance of SS2D. The SS2D performs selective scanning in four directions on the vision sequence.

Recently, SSM was introduced to the CFF. Peng et al. \cite{18} proposed a FusionMamba block, serving as a plug-and-play information fusion module. Xie et al. \cite{17} integrated the visual SMM with dynamic convolution and channel attention for image fusion. Li et al. \cite{21} proposed a coupled SSM for coupling state chains of multiple modalities. Zhou et al. \cite{16} performed feature enhancement for each modality using SSM before the feature fusion. However, the above methods ignored to construct the hidden state across different modalities. Thus, we proposed the CM-SS2D module to construct cross-modal vision sequence and obtains the hidden state of one modality through the other.

\section{PROPOSED METHOD}

\subsection{Overview} 
The overall architecture of CM-SSM is illustrated in Fig. 2. It features a typical encoder-decoder architecture, comprising two image encoders for extracting features from RGB and thermal images, four CM-SSA modules that perform RGB-T feature fusion across four stages, and an MLP decoder for predicting semantic segmentation maps. To balance the lightweight design with the performance of the model, we utilized EfficientVit-B1 \cite{22} for feature extraction. Both encoders are pretrained on the ImageNet-1K dataset with the resolution of 288 × 288 pixels. Given the input images $\mathbb{\emph{I}} \in \mathbb{R}^{3 \times \text{H} \times \text{W}}$, we obtained four scales of features $\left\{R_{1}, R_{2}, R_{3}, R_{4}\right\}$ and $\left\{T_{1}, T_{2}, T_{3}, T_{4}\right\}$ with the following dimensions (channel × height × width): 32 × H/4 × W/4, 64 × H/8 × W/8, 128 × H/16 × W/16, 256 × H/32 × W/32. In each encoder stage, a CM-SSA module is designed to integrate the local association obtained from convolution and the global association derived from the CM-SS2D, obtaining fused features $\left\{F_{1}, F_{2}, F_{3}, F_{4}\right\}$. The CM-SS2D constructs the cross-modal SSM between RGB-T features by constructing cross-modal visual sequence and generating hidden state of one modality from the other. Finally, an MLP  \cite{23} decoder with a hidden layer of 128 neurons is used to generate semantic segmentation maps.

\subsection{Cross-modal 2D-Selective-Scan (CM-SS2D)}

The core of the SMM in achieving global association lies in the design of the hidden state. In Mamba, the hidden state contains both the current input and previous state, gradually acquiring global memory during the selective scanning. When modifying the SSM to the CFF, the hidden state should simultaneously contain memory of the previous state and information from each modality.  However, previous studies \cite{17,18,34} have concentrated on utilizing multi-modal information to generate trainable parameters, overlooking the role of the hidden state. Thus, we proposed CM-SS2D to construct the cross-modal SSM from a novel perspective. As shown in Fig.  3, CM-SS2D consists of three steps: 1)cross-modal selective scanning, 2) cross-modal state space modeling, and 3) scan merging.

\emph{Cross-modal selective scanning}: Before constructing the SSM, we first build the cross-modal visual feature sequence.  Given the input RGB and thermal features $F_{\text{R}}$ and $F_{\text{T}} \in\mathbb{R}^{\text{C} \times \text{H} \times \text{W}}$, we perform cross-modal selective scan in four directions similar to the SS2D module. The difference lies in our approach of performing cross-modal scanning in the sequence of ‘RGB–thermal–RGB’, obtaining the cross-modal visual feature sequence $f_{\text{c}} = \{f_{\text{c}}^1, f_{\text{c}}^2, f_{\text{c}}^3, f_{\text{c}}^4\} \in \mathbb{R}^{4 \times \text{C} \times 2\text{HW}}$. As shown in Fig. \ref{figure3}, $f_{\text{c}}^1 = \{r_1, t_1, r_2, t_2, \ldots, r_{\text{k}}, t_{\text{k}}, \ldots, r_{\text{HW}}, t_{\text{HW}}\}$, where~$r_{\text{k}}$ and~$t_{\text{k}} \in \mathbb{R}^{\text{C}}$.

\emph{Cross-modal state space modeling}: We obtained the parameters \emph{B}, \emph{C} and $\Delta$ from $f_{\text{c}}$  through the linear layer and defined the learnable parameters \emph{A} and \emph{D}. The process of parameter initialization can be formulated as follows:
\begin{equation}
B, C = Linear(f_{\text{c}})
\end{equation}
\begin{equation}
\Delta = Linear(Linear(f_{\text{c}}))
\end{equation}
where the \emph{Linear}(.) denotes the linear full connection layer. Then we performed the discretization operation \cite{15} to obtain $\bar{A}$ and $\bar{B}$, which can be formulated as:
\begin{equation}
\begin{cases} 
\bar{A} = \exp(\Delta A) \\
\bar{B} = \Delta B 
\end{cases}
\label{eq:your_label}
\end{equation}

For the RGB modality, we build the hidden state $\bar{r}_\text{k}$ using the current input $r_\text{k}$ and the last state of thermal modality $\bar{t}_\text{k-1}$. The $\bar{r}_\text{k}$ contains cross-modal information and the memory of the previous states in the visual sequence. Then the output $r_{\text{k}}'$ are derived by combing $\bar{r}_{\text{k}}$ and $r_{\text{k}}$, further incorporating the influence of the current input. For the thermal modality, the hidden state $\bar{t}_{\text{k}}$ are derived using the current input $t_{\text{k}}$ and the last state of RGB modality $\bar{r}_{\text{k}-1}$, and the output $t_{\text{k}}'$ are derived by combing $\bar{t}_{\text{k}}$ and $t_{\text{k}}$. The above operation can be formulated as:

\begin{equation}
\begin{cases} 
\bar{r}_{\text{k}} = \bar{A} \bar{t}_{\text{k}-1} + \bar{B} r_{\text{k}} \\
r_{\text{k}}' = \bar{C} \bar{r}_{\text{k}} + \bar{D} r_{\text{k}} 
\end{cases}
\label{eq:eq4}
\end{equation}

\begin{equation}
\begin{cases} 
\bar{t}_{\text{k}} = \bar{A} \bar{r}_{\text{k}-1} + \bar{B} t_{\text{k}} \\
t_{\text{k}}' = \bar{C} \bar{t}_{\text{k}} + \bar{D} t_{\text{k}} 
\end{cases}
\label{eq:eq5}
\end{equation}

 \emph{Scan merging}: After the cross-modal state space modeling, we obtain the rectified RGB and thermal features from four directions, namely $f_{\text{r}}' = \{f_{\text{r}}^{1'}, f_{\text{r}}^{2'}, f_{\text{r}}^{3'}, f_{\text{r}}^{4'}\}$ and $f_{\text{t}}' = \{f_{\text{t}}^{1'}, f_{\text{t}}^{2'}, f_{\text{t}}^{3'}, f_{\text{t}}^{4'}\}$, where $f_{\text{r}}^{1'} = \{r_1', r_2', ..., r_\text{k}', ..., r_{\text{HW}}'\}$ and $f_{\text{t}}^{1'} = \{t_1', t_2', ..., t_\text{k}', ..., t_{\text{HW}}'\}$. Subsequently, we sum up each $r_\text{k}'$ and $t_\text{k}'$ in $f_{\text{r}}'$ and $f_{\text{t}}'$ to obtain $F_{\text{R}}'$ and $F_{\text{T}}' \in \mathbb{R}^{\text{C} \times \text{H} \times \text{W}}$, respectively. In this way, each position vector contains the embedding of all directions.
\subsection{Cross-modal State Space Association (CM-SSA)}

In addition to the global association capabilities of CM-SS2D, local association is also crucial for enhancing segmentation performance. Thus we proposed the CM-SSA to combine CM-SS2D and traditional convolution, as shown in Fig. 2. Firstly, the global association is achieved through CM-SS2D, with linear, activation and layer-normalization layers appended to enhance the fitting ability of the model. This can be formulated as: 
\begin{equation}
\begin{cases} 
F_{\text{R}} = SiLU(DWConv(Linear(R_{\text{i}}))) \\
F_{\text{T}} = SiLU(DWConv(Linear(T_{\text{i}}))) 
\end{cases}
\label{eq:eq6}
\end{equation}
\begin{equation}
\begin{cases} 
F_\text{R}' = CMSS2D(F_\text{R}) \\
F_\text{T}' = CMSS2D(F_\text{T}) 
\end{cases}
\label{eq:eq7}
\end{equation}
where $R_{\text{i}}$ and $T_{\text{i}}$ denote the RGB and thermal featuresobtained from feature extraction. $DWConv(.)$ and $SiLU(.)$ denote the depth-wise separated convolution and SiLU activation, respectively. $CMSS2D(.)$ represents the proposed CM-SS2D module. The global association features $G_{\text{R}}$ and $G_{\text{T}}$ are then derived from residual connection, which can be formulated as:
\begin{equation}
\begin{cases} 
G_{\text{R}} = R_{\text{i}} + Linear(LN(F_\text{R}')) \otimes SiLU(Linear(R_{\text{i}})) \\
G_{\text{T}} = T_{\text{i}} + Linear(LN(F_\text{T}')) \otimes SiLU(Linear(T_{\text{i}})) 
\end{cases}
\label{eq:eq8}
\end{equation}
where \emph{LN}(.) and $\otimes$ denotes the layer-normalization and addition operations, respectively. In addition, the local association features \emph{L} is obtained from the convolution, which can be formulated as:
\begin{equation}
L = CBR_{3 \times 3}(Cat(R_{\text{i}}, T_{\text{i}}))
\label{eq:eq9}
\end{equation}
where $CBR_{3 \times 3}(.)$ denotes the convolution with the kernel size of 3, followed by the batch-normalization and ReLU activation. $Cat(.)$ denotes the concatenation operation. Finally, we combine $G_{\text{R}}$, $G_{\text{T}}$ and $L$ to incorporate both local and global associations. This can be formulated as:
\begin{equation}
F_{\text{i}} = CBR_{1 \times 1}(Cat(R_{\text{i}}, T_{\text{i}}, L))
\label{eq:eq10}
\end{equation}
where $CBR_{1\times1}$(.) denotes the convolution with the kernel size of 1, followed by the batch-normalization and ReLU activation.

\subsection{Loss Function}
With fused features $F_{\text{i}}$, an MLP decoder is utilized to predict the semantic segmentation maps. Considering the varying pixel distribution of each object in the samples, we utilize the weighted cross-entropy loss. In addition, dice loss was utilized to further emphasize the segmentation of the foreground. The loss function can be formulated as follows:
\begin{equation}
Loss1 = L_{\text{ce}}(Pre, GT) + L_{\text{dice}}(Pre, GT)
\label{eq:eq11}
\end{equation}
where $Pre$ and $GT$ denote the prediction and the ground truth (GT) images, respectively.
\begin{table*}[t]
\caption{PERFORMANCE COMPARISON WITH SOTA METHODS ON THE CART DATASET. THE BEST AND SECOND-BEST RESULTS ARE HIGHLIGHTED IN {\color[HTML]{CB0000}RED} AND {\color[HTML]{3166FF}BLUE} IN EACH COLUMN, RESPECTIVELY.}
\label{table_example}
\begin{center}
\renewcommand{\arraystretch}{1.25}
\begin{tabular}{ccccccccccccc}
\hline
\multicolumn{2}{c}{}                                 & \multicolumn{10}{c}{IoU↑}                                                                                                                                                                                                                                                                                                                                                                                    &                              \\ \cline{3-12}
\multirow{-2}{*}{Model} & \multirow{-2}{*}{Backbone} & \begin{tabular}[c]{@{}c@{}}Bare \\ Ground\end{tabular} & \begin{tabular}[c]{@{}c@{}}Rocky \\ Terrain\end{tabular} & \begin{tabular}[c]{@{}c@{}}Developed \\ Structures\end{tabular} & Road                         & Shrubs                       & Trees                        & Sky                          & Water                        & Vehicles                     & Person                       & \multirow{-2}{*}{mIoU↑}      \\ 
\hline
GMNet$_{21}$&ResNet152&81.5&88.8&83.8&60.3&70.8&79.5&92.0&97.4&54.8&17.8&72.7\\
EGFNet$_{22}$&	ResNet101&	80.3&	85.2&	56.3&	60.5&	69.9&	78.4&	92.8	&97.4	&57.2	&17.1	&69.5\\
EAEFNet$_{23}$&	ResNet152&	80.5&	86.3&	81.3&	54.5&	70.2&	\color[HTML]{CB0000}80.2&	\color[HTML]{CB0000}93.7&	\color[HTML]{3166FF}97.6&	48.1&	0.0&	69.2\\
SGFNet$_{23}$&	ResNet152&	77.0&	87.8&	63.7&	55.6&	68.5&	76.5&	92.1&	96.7&	48.0&	18.0&	68.5\\
SFAF-MA$_{23}$&	ResNet152&	75.1&	77.2&	66.9&	47.3&	65.1&	71.4&	90.7&	96.3&	38.8&	2.5&	63.1\\
ECM$_{23}$&	ResNet152&	67.9&	86.1&	46.6&	59.1&	67.0&	75.0&	87.6&	92.2&	57.7&	9.3&	64.9\\
CMNeXt$_{23}$&	Mit-B2&	80.1&	87.8&	84.1&	58.6&	69.9&	78.5&	92.5&	97.3&	54.3&	17.8&	72.1\\
CMX$_{23}$&	Mit-B2&	\color[HTML]{3166FF}82.4&	\color[HTML]{CB0000}90.3&	\color[HTML]{3166FF}85.0&	60.4&	\color[HTML]{3166FF}71.0&	\color[HTML]{3166FF}79.9&	\color[HTML]{3166FF}93.6&	\color[HTML]{3166FF}97.6&	57.9&	22.2&	\color[HTML]{3166FF}74.0\\
CRM$_{24}$& 	Swin-B&	81.6&	\color[HTML]{3166FF}90.0&	\color[HTML]{CB0000}85.1&	\color[HTML]{CB0000}61.6&	69.5&	77.9&	92.3&	97.4&	37.7&	20.6&	71.4\\
CLNet-T$_{24}$&	Mit-B4&	81.8&	89.3&	84.8&	\color[HTML]{3166FF}61.5&	\color[HTML]{CB0000}71.3&	79.5&	93.4&	97.5&	52.3&	\color[HTML]{3166FF}25.1&	73.7\\
CAINet$_{24}$&	MobileNetV2&	80.2&	88.4&	78.4&	57.1&	68.8&	77.0&	92.1&	96.9&	43.9&	17.2&	70.0\\
MCNet-T$_{25}$&	ConvNeXtV2-B&	80.5&	86.9&	83.7&	58.0&	69.7&	78.0&	92.9&	\color[HTML]{CB0000}97.7&	\color[HTML]{3166FF}58.1&	\color[HTML]{CB0000}31.5&	73.7\\
\hline
CM-SSM                 & EfficientVit-B1             & {\color[HTML]{CB0000} 82.6}                           & 89.7                                                    & 84.9                                    & 61.0 & {\color[HTML]{CB0000}71.3}                        & 79.1                        & 92.9                        & {\color[HTML]{3166FF} 97.6} & {\color[HTML]{CB0000} 62.0} & 24.5 & {\color[HTML]{CB0000} 74.6} \\ 
\hline
\end{tabular}
\end{center}
\end{table*}

\begin{table*}[t]
\caption{PERFORMANCE COMPARISON WITH SOTA METHODS ON THE PST900 DATASET.}
\label{table_example}
\begin{center}
\renewcommand{\arraystretch}{1.25}
\begin{tabular}{ccccccc}
\hline
                        &                            & \multicolumn{4}{c}{IoU↑}                                                                                                                                     &                              \\ \cline{3-6}
\multirow{-2}{*}{Model} & \multirow{-2}{*}{Backbone}                & Hand-Drill                   & Backpack                     & Fire-Extinguisher            & Survivor                     & \multirow{-2}{*}{mIoU↑}\\ \hline
MMSMCNet$_{23}$&	Mit-B3&	    62.4&	\color[HTML]{3166FF}89.2&	73.3&	74.7&	79.8\\
SGFNet$_{23}$&	ResNet152&	76.7&	85.4&	75.6&	76.7&	82.8\\
DBCNet$_{23}$&	ResNet101&	\color[HTML]{3166FF}77.2&	82.7&	73.0&	76.7&	81.8\\
CLNet-T$_{24}$&	Mit-B4&	    68.5&	87.9&	72.7&	75.7&	80.8\\
MDNet$_{25}$&	    Mit-B2&	    69.5&	\color[HTML]{CB0000}89.4&	76.8&	\color[HTML]{CB0000}80.2&	\color[HTML]{3166FF}83.0\\
C$^{4}$Net-Mit$_{25}$ &	Mit-B2&	    69.4&	88.4&	\color[HTML]{3166FF}79.6&	\color[HTML]{3166FF}79.6&	82.6\\
\hline
{CM-SSM}&EfficientVit-B1&	\color[HTML]{CB0000}80.5&	85.4&	\color[HTML]{CB0000}85.7&	78.1&	\color[HTML]{CB0000}85.9\\
\hline
\end{tabular}
\end{center}
\end{table*}

\section{EXPERIMENT RESULTS}

\subsection{Datasets and Evaluation Metrics}

We employed the publicly achieved CART and PST900 datasets for the training, evaluation and testing of the proposed CM-SSM.

\textbf{The CART dataset} \cite{1} contains 2282 pairs of aligned RGB-T images captured in various terrains, including rivers, lakes, coastlines, deserts and forests. It provides semantic segmentation labels for 10 classes: bare ground, rocky terrain, developed structures, road, shrubs, trees, sky, water, vehicles, and person. Both the RGB and thermal images have a spatial resolution of 960 × 640 pixels. The dateset was randomly partitioned into train/val/test set at a 6:1:1 ratio.

\textbf{The PST900 dataset} \cite{20} comprises 894 pairs RGB-T images captured in challenging underground environments. It provides semantic segmentation annotations for 4 classes: hand drill, backpack, fire extinguisher and survivor. Both the RGB and thermal images have a spatial resolution of 1280 × 
720 pixels. The datasets was divided into train/test set with at a ratio of 2:1.

We utilized the mean intersection over union (mIoU) to assess the segmentation performance. Besides, we employed FLOPs, Params, and FPS to access the complexity of the model.

\subsection{Implementation Details}

The CM-SSM was trained, evaluated and tested using the Pytorch library, and all experiments were conducted on an RTX 4090 GPU. Before training, the EfficientVit-B1 backbone was initialized with the pretrained weights from ImageNet-1K (with the resolution of 288 × 288 pixels), whereas the remaining part was initialized randomly. Image resizing, random cropping and flipping were used to augment the data during training, as the configure in the original benchmark \cite{1, 20}. The training was optimized using Ranger with  a weight decay of $5\times10^{-4}$, with a training batch of 8. The initial learning rate was set to $1\times10^{-4}$ and multiplied by $(1-\frac{iter}{max\_iter})^{power}$ during training, with a power of 0.9. We set the training epoch to 300 and selected the best weight based on the validation set. When testing and training the PST900 dataset, the batch size, and training epoch were changed to 2 and 200, respectively.

\begin{figure*}[t]
      \centering
      \includegraphics[width=\textwidth]{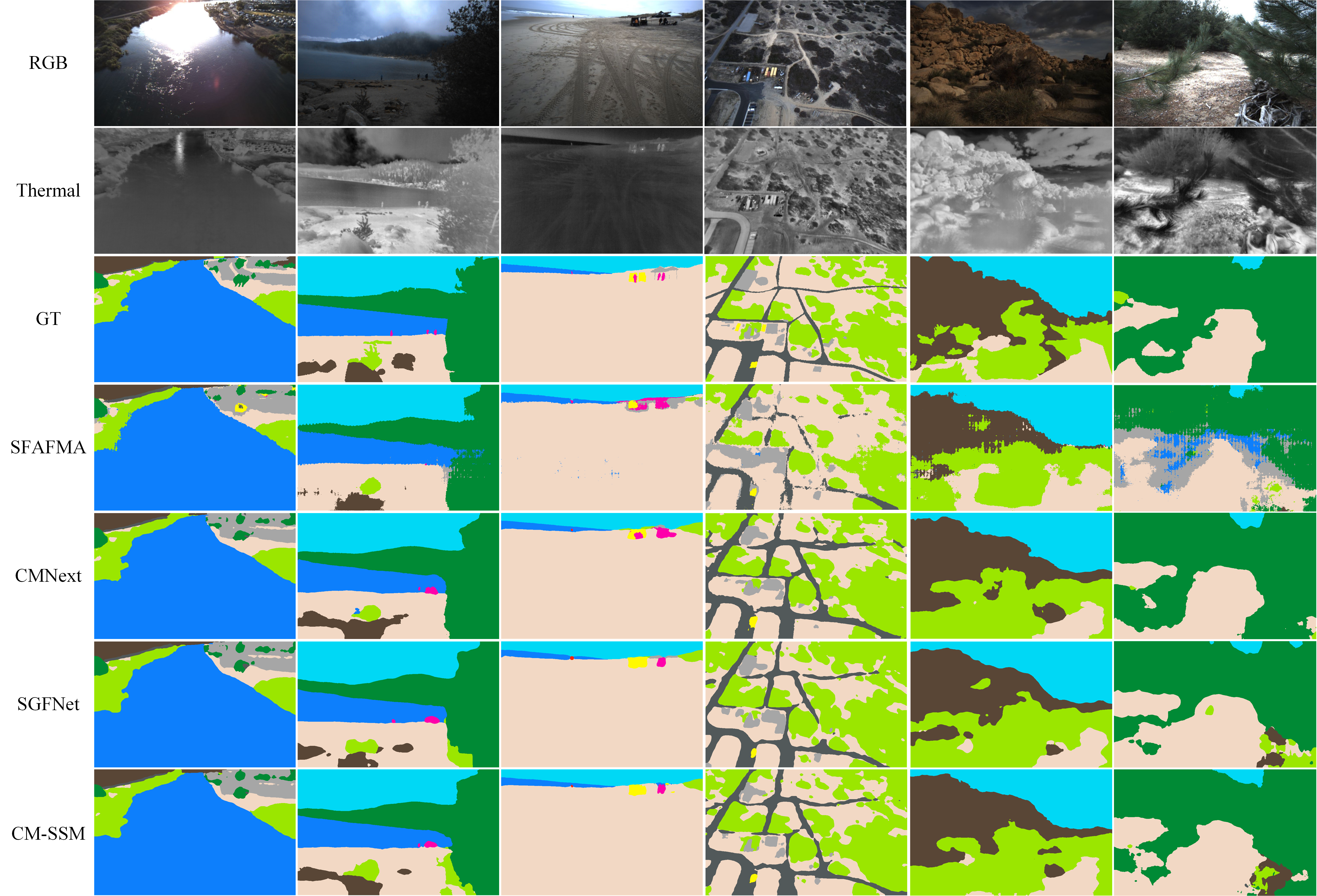}
      \caption{Visual diagram of the segmentation maps of CM-SSM and some SOTA models on the CART dataset.}
      \label{figure4}
\end{figure*}
\begin{table}[h]
\caption{COMPLEXITY COMPARISON ON AN RTX 4090 GPU, WITH THE IMAGE RESOLUTION OF 480 × 640 PIXLES.}
\label{table_example}
\begin{center}
\setlength{\tabcolsep}{1.35mm}{
\renewcommand{\arraystretch}{1.25}
\begin{tabular}{cccccc}
\hline
Model     & Backbone        & FLOPs(G)↓      & Params(M))↓      & FPS↑            & mIoU↑          \\ \hline
SGFNet$_{23}$  & ResNet152       & 143.73         & 125.25         & 20          & 68.5          \\
CMNeXt$_{23}$  & Mit-B2          & 68.70          & 58.68          & 67          & 72.1          \\
CMX$_{23}$     & Mit-B2          & 67.20          & 66.57          & 63          & 74.0          \\
CLNet-T$_{24}$ & Mit-B4          & 217.85         & 130.84         & 28          & 73.7          \\
CAINet$_{24}$&	    MobileNetV2&123.62&	\textbf{12.16}&	61& 70.0\\
MCNet-T$_{25}$&	ConvNeXt-B&	278.85&	199.79&	32 & 73.7\\
\hline
CM-SSM    & EfficientVit-B1 & \textbf{10.34} & 12.59 & \textbf{114} & \textbf{74.6} \\ \hline
\end{tabular}}
\end{center}
\end{table}
\subsection{Comparison with SOTA models}
\textbf{1)	CART dataset.} On the CART dataset, the CM-SMM was compared to 12 SOTA models: GMNet \cite{3}, EAEFNet \cite{24}, CMNeXt \cite{6}, CMX \cite{5}, SGFNet \cite{25}, EGFNet \cite{26}, SFAF-MA \cite{27}, ECM \cite{28}, CRM \cite{29}, CLNet-T \cite{2}, CAINet \cite{30}, MCNet-T \cite{36}. Table $\mathrm{I}$ presents a quantitative evaluation of the aforementioned models, including the performance for 10 classes and an overall assessment. Overall, the CM-SSM achieved the best performance with an mIoU of 74.6\%. Despite only surpassing CMX—the second-best model—by 0.6\% in mIoU, CM-SSM demonstrates remarkable efficiency, reducing FLOPs and parameters to one-sixth and doubling the inference speed, as presented in Table $\mathrm{III}$. Compared with the lightweight model CAINet, which has a similar number of parameters, CM-SSM improves mIoU by 4.6\% and achieves higher IoU across all 10 categories. For more challenging categories such as vehicle and person, CM-SSM improves IoU by 18.1\% and 7.3\%, respectively, demonstrating its significant advantage among lightweight models. Fig. 4 displays 6 sets of images for a intuitive assessment. Compared with SOTA models, the segmentation maps of CM-SMM is the closet to the GT images, exhibiting higher accuracy at the regions of small targets. 

\textbf{2)	PST900 dataset.} On the PST900 dataset, the CM-SMM was compared to 6 SOTA models: MMSMCNet \cite{8}, SGFNet \cite{25}, DBCNet \cite{31},C$^{4}$Net-Mit \cite{32}, CLNet-T \cite{2} and MDNet \cite{16}. Table $\mathrm{II}$ presents a quantitative evaluation of the aforementioned methods. The CM-SSM achieved the best performance with an mIoU of 85.9\%. In addition, CM-SSM achieved IoU improvements of 11.0\% and 8.9\% over MDNet on the hand-drill and fire-extinguisher categories, respectively. This demonstrates the CM-SSM generalized well on the PST900 dataset.

\textbf{3)	Complexity analysis.} To evaluate the practicality of the model on resource-constrained filed robots, we performed a complexity assessment. Table $\mathrm{III}$ presents the FLOPs, Params, and FPS of the CM-SSM compared with four SOTA models. While achieving optimal performance, the CM-SSM maintains a lower inference burden and fewer parameters. In addition, it achieved an inference speed of 114 FPS, demonstrating its real-time capability.

\begin{table}[t]
\caption{ABLATION STUDIES OF THE CM-SS2D AND CM-SSA.}
\label{table_example}
\begin{center}
\renewcommand{\arraystretch}{1.25}
\begin{tabular}{cccc}
\hline
Variations       &  Backbone               & Fusion Strategy & mIoU↑           \\ 
\hline
1&\multirow{7}{*}{EfficientVit-B1} &Addition    & 72.6          \\
2                                 &&FFM             & 73.4           \\
3                                 &&CDA             & 74.0           \\
4                                 &&MDFusion        & 73.0          \\
5                                 &&CroMB+ConMB     & 73.5          \\
6                                 &&w/o CM-SS2D     & 73.3          \\
7                                 & &CM-SSA          & \textbf{74.6} \\ \hline
\end{tabular}
\end{center}
\end{table}

\subsection{Ablation Studies}  
To prove the effectiveness of the proposed CM-SS2D and CM-SSA modules, we conducted a series of ablation studies. Table $\mathrm{IV}$ presents several variations of the model. We utilized the EfficientVit-B1 as the backbone for all variations, and modified the CFF strategies. As a baseline variation, we used the addition operation to fuse RGB-T features, which introduced no additional parameters. In the second variation ‘FFM’, we utilized the fusion strategy in the CMX \cite{5}, which is designed based on the Transformer-based attention utilized in the channel dimension. The third variation ‘CDA’ adopts the fusion strategy from MCNet-T \cite{36}, which combines transformer across both spatial and channel dimensions. The fourth variation ‘MDFusion’ is introduced based on the MDFusion \cite{16}. It uses SS2D for further feature extraction, without the cross-modal SSM. The fifth variation is based on the Cross Mamba and Concat Mamba blocks \cite{34}, which obtain the state space parameters of one modality from the other. The sixth variation removes the CM-SS2D module from CM-SSA to verify its effectiveness. The last variation is the original CM-SSM.

Compared to the baseline variation, our CM-SSM showed the best improvement, with a $2.0\%$ increase in mIoU. This demonstrates the effectiveness of the proposed CM-SSA module. Besides, it surpassed the variations with ‘FFM’ and ‘CDA’ by $1.2\%$ and 0.6\% in mIoU, respectively, demonstrating that it outperformed fusion strategies based on the Transformer-based attention. Our approach demonstrated superior performance compared to both ‘MDFusion’ and ‘CroMB+ConMB’, further validating our core hypothesis that the construction of cross-modal hidden states is pivotal to advancing cross-modal SSM. When comparing the sixth variation with CM-SSM, we observed a $1.3\%$ decrease in mIoU. This confirms the necessity of the CM-SS2D module, which constructs the cross-modal SSM. 

\section{CONCLUSION}

We proposed a real-time CM-SSM for filed robots working in wild scenes, improving the accuracy of RGB-T segmentation. The CM-SSM contains a CM-SS2D module to construct the corss-modal visual sequence and derive hidden state of one modality from the other, and a CM-SSA module to incorporate the local and global information. With modules above, CM-SSM establishes global correlation across various modalities while keeps linear computational complexity with respect to image resolution. It achieved an inference speed of 114 frames per second on an RTX 4090 GPU, with 12.59M parameters. Extensive experiments conducted on the CART and PST900 datasets demonstrated its competitive performance compared with SOTA models. Future work will explore the use of knowledge distillation to enhance the generalization ability of lightweight models across diverse terrains.






\bibliographystyle{IEEEtran}
\bibliography{IEEEabrv,IEEEexample}

\end{document}